\documentclass[letterpaper, 10 pt, conference]{ieeeconf}  

\IEEEoverridecommandlockouts                              

\usepackage{cite}
\usepackage{amsmath,amssymb,amsfonts}
\usepackage{algorithmic}
\usepackage{graphicx}
\usepackage{float}
\usepackage{textcomp}
\usepackage{xcolor}
\usepackage{amsfonts,amssymb}
\usepackage{verbatim} 
\usepackage{stfloats}
\usepackage{multirow}
\usepackage{microtype}
\usepackage{optidef}
\usepackage{hyperref}
\usepackage{bm}
\usepackage{multirow}
\usepackage{booktabs}
\usepackage{bigstrut}
\usepackage{hyperref}
\usepackage{threeparttable}
\usepackage{makecell}
\usepackage{algorithm2e} 
\usepackage{adjustbox}
\usepackage[english]{babel}
\usepackage{color, colortbl}
\definecolor{Gray}{gray}{0.95}

\title{\LARGE \bf
V2X-Lead: LiDAR-based End-to-End Autonomous Driving \\with Vehicle-to-Everything Communication Integration
}

\author{Zhiyun Deng$^{1}$, Yanjun Shi$^{2}$, and Weiming Shen$^{1}$
\thanks{This work was supported in part by the National Key Research and Development Program of China under Grant 2018YFE0197700, and the Fundamental Research Funds for the Central Universities of China under Grants 2021yjsCXCY048 and 2021GCRC058.}
\thanks{$^{1}$Z. Deng and W. Shen are with the School of Mechanical Science and Engineering, Huazhong University of Science and Technology, Wuhan, 430074, China.
        {\tt\small dengzy@hust.edu.cn, \tt\small shenwm@hust.edu.cn}}%
\thanks{$^{2}$Y. Shi is with the School of Mechanical Engineering, Dalian University of Technology, Dalian, 116024, China.
        {\tt\small syj@dlut.edu.cn}}
}

\begin{document}

\maketitle
\thispagestyle{empty}
\pagestyle{empty}

\begin{abstract}
This paper presents a LiDAR-based end-to-end autonomous driving method with Vehicle-to-Everything (V2X) communication integration, termed V2X-Lead, to address the challenges of navigating unregulated urban scenarios under mixed-autonomy traffic conditions. 
The proposed method aims to handle imperfect partial observations by fusing the onboard LiDAR sensor and V2X communication data. 
A model-free and off-policy deep reinforcement learning (DRL) algorithm is employed to train the driving agent, which incorporates a carefully designed reward function and multi-task learning technique to enhance generalization across diverse driving tasks and scenarios.
Experimental results demonstrate the effectiveness of the proposed approach in improving safety and efficiency in the task of traversing unsignalized intersections in mixed-autonomy traffic, and its generalizability to previously unseen scenarios, such as roundabouts. 
The integration of V2X communication offers a significant data source for autonomous vehicles (AVs) to perceive their surroundings beyond onboard sensors, resulting in a more accurate and comprehensive perception of the driving environment and more safe and robust driving behavior.
\end{abstract}

\section{Introduction}
\begin{figure}[t]
\centering
\includegraphics[width=0.48\textwidth]{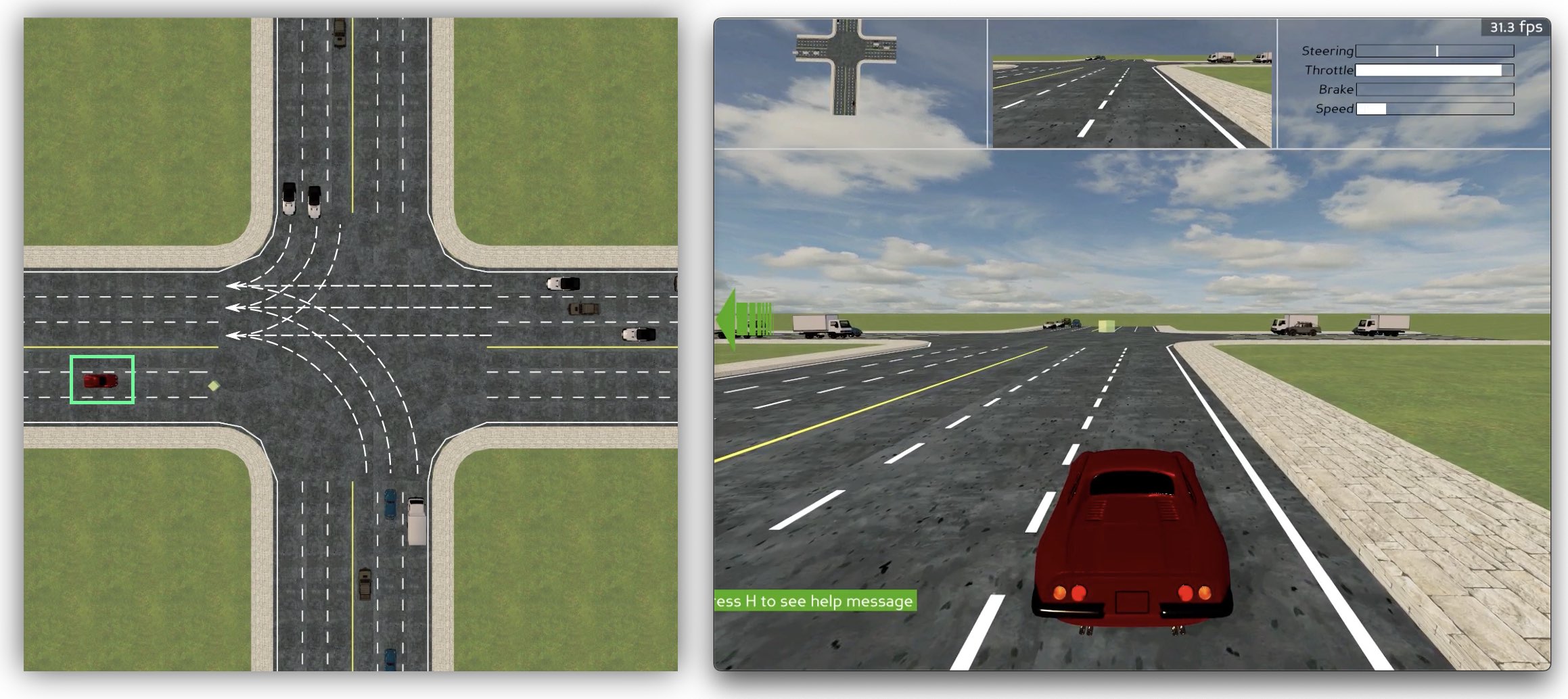}
\caption{Illustration of the driving scenarios in the mixed-autonomy traffic environment, where the AV colored in red is driven by an agent trained by reinforcement learning and is required to pass through an unsignalized intersection. In that case, the other vehicles are human-driven vehicles and are actuated by the Intelligent Driver Model (IDM).
}
\label{scenarios-intersection}
\end{figure}

Autonomous driving has made remarkable progress in recent years, positioning it as a potential catalyst for revolutionizing the transportation system. However, the real-world deployment of AVs poses significant challenges that must be addressed to ensure their safety. Among these challenges, navigating unregulated urban scenarios, particularly unsignalized intersections shown in Fig. \ref{scenarios-intersection}, under mixed-autonomy traffic conditions, stands out as a crucial issue \cite{khaitan2022state}.

Traditional optimization-based decision-making \cite{deng2023cooperative} and motion planning \cite{deng2022coevolutionary} techniques have limitations in complex environments with mixed-autonomy traffic due to the unpredictable behavioral patterns of human-driven vehicles (HDVs)\cite{wu2021flow,mu2022cooperative}. Furthermore, their dependence on pre-defined rules and a priori knowledge makes them less adaptable to new scenarios, limiting their generalization performances. Consequently, machine learning-based approaches, such as end-to-end driving \cite{tampuu2020survey}, have become popular in the research community. These approaches aim to learn driving policies from large amounts of data without relying on hand-crafted rules, and use deep neural networks to directly map raw sensory data, such as LiDAR point clouds, to driving control commands. This trend is gaining traction as a promising solution to the challenges of navigating AVs through unsignalized intersections.

However, end-to-end autonomous driving still faces limitations in perception and decision-making. Onboard sensors, such as cameras and LiDAR, have limited range and can be affected by adverse weather conditions, obstructed views, and occlusions. This can result in an incomplete or inaccurate perception of the driving environment, compromising the safety and efficiency of autonomous driving. Furthermore, AVs have limited access to real-time information about other vehicles, pedestrians, and infrastructure in their surroundings, making it difficult for AVs to predict and respond to potential hazards, such as sudden stops and turning vehicles. As a result, the decision-making process of AVs may be slower or less precise, leading to decreased safety and efficiency.

Fortunately, the emergence of V2X communication has provided an additional approach for AVs to perceive their environment beyond onboard sensors. By integrating V2X communication, AVs can communicate with other vehicles and infrastructure in their surroundings, providing a more accurate and comprehensive understanding of the driving environment. However, the integration of V2X communication brings a new challenge of sensor fusion, which needs to be addressed to ensure its effectiveness and suitability for real-life deployment.

\linespread{1.1}

Compared to existing studies, the key contributions of this paper are summarized as follows: 
1) We propose an innovative LiDAR-based end-to-end autonomous driving method with V2X communication integration, termed V2X-Lead. The proposed method can handle imperfect partial observations by fusing the onboard LiDAR sensor and V2X communication data. 
2) We present a model-free and off-policy DRL algorithm that employs a well-designed reward function to train a driving agent. Additionally, we utilize a multi-task learning technique to improve the generalization ability of the algorithm across different driving tasks and scenarios. 
3) We demonstrate the effectiveness of the proposed method in improving safety and efficiency in the task of traversing unsignalized intersections in mixed-autonomy traffic, and further provide evidence of the generalizability of the proposed method to previously unseen environments, such as roundabouts, through comparison experiments against benchmarking methods.

The remainder of this paper is organized as follows: 
Section \uppercase\expandafter{\romannumeral2} discusses some recent works in related areas. 
Section \uppercase\expandafter{\romannumeral3} presents a detailed description of the proposed methodology, followed by the experiment results and discussion in Section \uppercase\expandafter{\romannumeral4}.
Section \uppercase\expandafter{\romannumeral5} concludes this paper and discusses some open issues and future work.

\section{Related Work}
\subsection{End-to-End Autonomous Driving}
The research on autonomous driving can be classified into two categories: modular and end-to-end. 
The modular approach involves utilizing interconnected but self-contained modules, such as perception \cite{xu2022v2x}, localization \cite{li2021robust}, planning\cite{qureshi2019motion}, and control\cite{cui2020nonlinear}. These modules provide interpretability and allow for identifying the faulty module in case of unexpected behavior. 
However, building and maintaining such pipelines can be costly, and despite extensive research, the modular approach is still far from achieving complete autonomy.

End-to-end driving is an alternative approach to modular driving and has become a growing trend in autonomous driving research \cite{wu2019end,amini2018variational,chen2020end}. In this approach, the entire driving pipeline, from processing sensory inputs to generating steering and acceleration commands, is optimized as a single machine learning task. 
The driving model is either learned in a supervised fashion via imitation learning (IL) \cite{le2022survey} to mimic human drivers or through exploration and improvement of driving policy from scratch via reinforcement learning (RL) \cite{kiran2021deep, yan2022unified}. 
While IL often suffers from insufficient exposure to diverse driving situations during training, RL is more immune to this problem since it relies on online interactions with the environment.

\subsection{LiDAR-based Autonomous Driving}
Although humans can drive efficiently using solely visual cues, RL agents require substantial memory and computational resources to process images \cite{liu2021efficient, liang2018cirl}.
Moreover, there will be great differences between the images obtained from the real world and traffic simulators, which may result in the policy trained in simulators not being deployed or achieving the same performance in real life.
Alternatively, LiDAR can be another notable source of input for autonomous driving. LiDAR point clouds are insensitive to illumination conditions and can provide reasonable distance estimations.
The point cloud data derived from LiDAR can be transformed into either an occupancy grid map \cite{cai2021carl} or a polar grid map \cite{zhang2020polarnet}. Following this transformation, the data can be further visualized as a bird's-eye view map using the MotionNet method \cite{wu2020motionnet}.

\subsection{V2X Network for Autonomous Driving}
The V2X network enables real-time communication among vehicles, allowing vehicles to access a comprehensive set of information that cannot be obtained by using LiDAR or vision sensors alone. For instance, AVs can obtain information on traffic signals \cite{ferreira2011impact} and detailed states of surrounding vehicles \cite{deng2020conflict} even under adverse weather conditions. As a result, AVs equipped with V2X communicators can better understand their environment and respond quickly to potential hazards, thus offering significant benefits for accident avoidance \cite{deng2019cooperative} and mobility improvement \cite{deng2022longitudinal}. 
However, the benefits of V2X for autonomous driving have not been thoroughly investigated and leveraged. This motivates us to develop a safe and efficient autonomous driving algorithm with V2X communication integration and further pave the way for the applications of V2X-assisted autonomous driving at occlusion environments, such as unsignalized intersections.

\section{Methodology}
\linespread{1.1}
\begin{figure*}[t]
\centering
\includegraphics[width=0.95\textwidth]{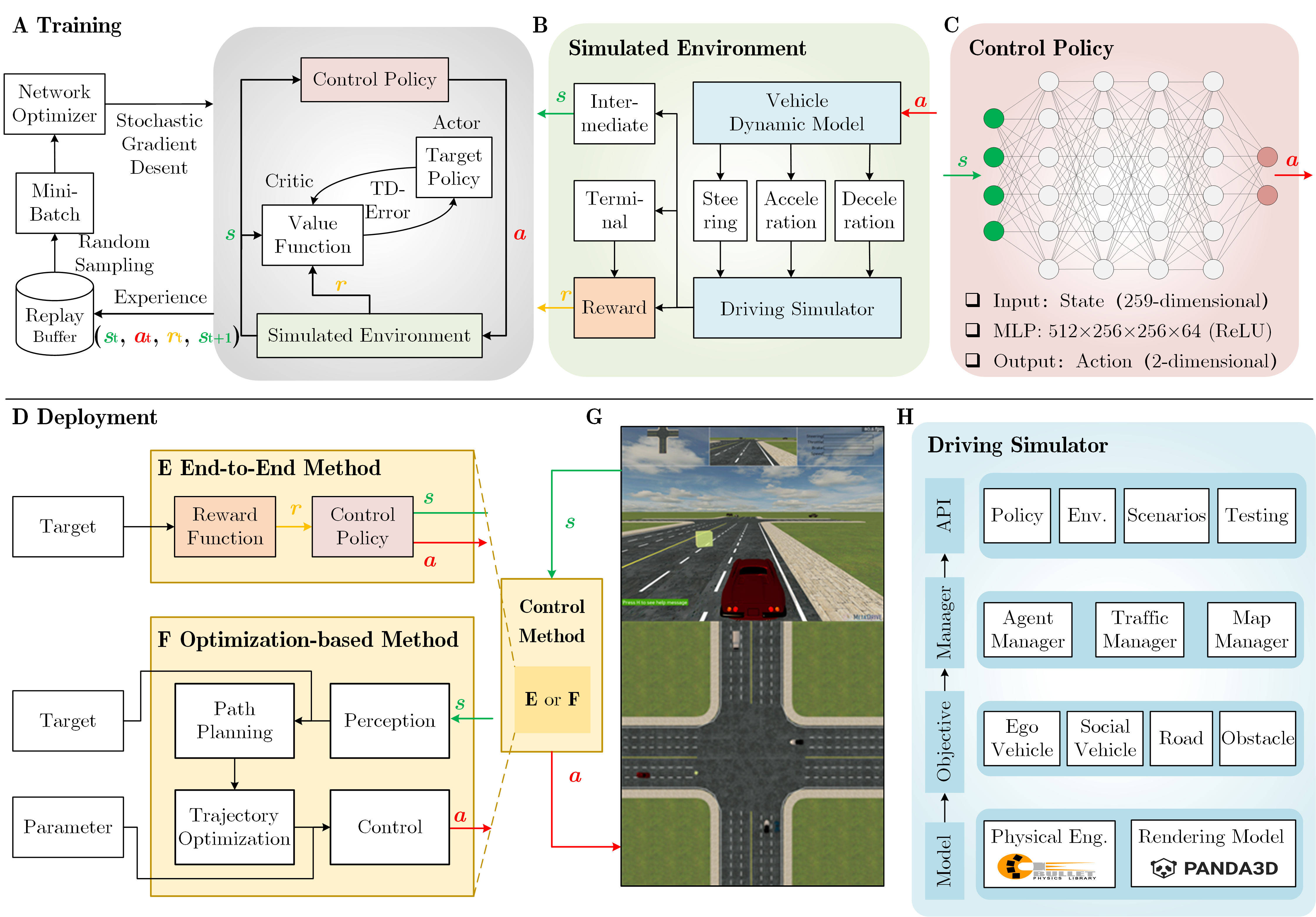}
\caption{Representation of the components of the proposed end-to-end driving architecture.
(A) Depiction of the learning loop. The control policy send control commands on the basis of the current state of the simulated environment and control targets. These data are sent to the priority-based replay buffer, which feeds data to the learner to update the policy.
(B) Our environment interaction loop, consisting of a vehicle dynamic model, driving simulator, and reward computation components.
(C) Our control policy is an MLP with four hidden layers that takes observation of environment and control outputs control commands.
(D-F) The control system implemented using either a conventional controller \cite{deng2023cooperative} composed of many subcomponents (F) or our end-to-end architecture using a single deep neural network to control the target vehicle directly (E).
(G-H) Illustration of the driving sumulator.
}
\label{result-architecture}
\end{figure*}
\subsection{Deep Reinforcement Learning}
DRL problems can be expressed as a system consisting of an agent and an environment \cite{graesser2019foundations}. The environment produces information that describes the state $s$ of the system, while the agent interacts with the environment by observing the state and then selecting an action $a$ to perform. Subsequently, the environment accepts the action and transitions into the next state while returning a reward $r$ to the agent. At each time $t$, the cycle of $(\text{state}\rightarrow \text{action}\rightarrow \text{reward})$ is repeated until reaching either a terminal state or a maximum time step $t=T$.

The state and action at time $t$ can be formally described as $s_t\in \mathcal{S}$ and $a_t\in \mathcal{A}$, where $\mathcal{S}$ and $\mathcal{A}$ denote the set of all possible states in an environment and possible actions defined by an environment, respectively. Additionally, the rewards can be calculated by $r_t= \mathcal{R}(s_t,a_t,s_{t+1})$, where $\mathcal{R}$ denotes the reward function that assigns a positive, negative, or zero scalar to each transition $(s_t,a_t,s_{t+1})$. 
After that, the transition function that determines how an environment transitions from one state to the next is formulated as a Markov Decision Process (MDP) under the assumption that the transition to the next state $s_{t+1}$ only depends on the previous state $s_t$ and action $a_t$. In other words, the next state is sampled from a probability distribution $P(s_{t+1}|s_t,a_t)$, i.e.,
\begin{equation}\label{backgroud1}
s_{t+1}\sim P(s_{t+1}|s_t,a_t).
\end{equation}

Having defined these notations, the MDP formulation of a general DRL problem can be presented in a 4-tuple form, i.e.,  $\mathcal{M}=(\mathcal{S},\mathcal{A},P,\mathcal{R})$. However, the model-free RL problem assumes that agents do not have access to the underlying transition function or reward function. Therefore, the only way an agent can get information about these functions is through the states, actions, and rewards it actually experiences in the environment\textemdash that is, the tuples $(s_t,a_t,r_t)$. Moreover, a sequence of experiences over an episode is referred to as a trajectory, i.e., $\tau=(s_0,a_0,r_0),(s_1,a_1,r_1),\dots,(s_T,a_T,r_T)$. Then, the overall reward $R(\tau)$ can be defined as a discounted sum of the instant rewards in a trajectory, i.e.,
\begin{equation}\label{backgroud-reward}
R(\tau) = \sum^T_{t=0}\gamma^t\,r_t,
\end{equation}
\noindent where $\gamma\in[0,1]$ is the discount factor that changes the way future rewards are valued. Besides, the objective $J(\tau)$ is simply the expectation of the returns over many trajectories, i.e.,
\begin{equation}\label{backgroud-objective}
J(\tau) = \mathbb{E}_{\tau\sim\pi}\left[R(\tau)\right] = \mathbb{E}_\tau\left[\sum^T_{t=0}\gamma^t\,r_t \right].
\end{equation}

The target of RL is to learn a policy denoted by $\pi$, which maps state to action: $a\sim \pi($s$)$. In other words, the policy determines how an agent produces actions in the environment to maximize the long-term discounted cumulative reward, i.e., 
\begin{equation}\label{backgroud-pi}
\pi = \arg\max_\pi \;\mathbb{E}_{\tau\sim\pi}\left[\sum^T_{t=0}\gamma^t\,r_t\right].
\end{equation}

In order to help an agent understand how good the states and available actions are in terms of the expected future return, the value function $V^\pi$ and $Q$-value function $Q^\pi$ are defined as
\begin{equation}\label{backgroud-V}
V^\pi(s_t)=\mathbb{E}_{s_0=s_t,\tau\sim\pi}\left[ \sum^T_{t=0}\gamma^t\,r_t \right],
\end{equation}

\begin{equation}\label{backgroud-Q}
Q^\pi(s_t,a_t)=\mathbb{E}_{s_0=s_t,a_0=a_t,\tau\sim\pi}\left[ \sum^T_{t=0}\gamma^t\,r_t \right],
\end{equation}
\noindent Here, with the assumption that the agent continues to act according to its current policy $\pi$, $V^\pi$ measures the expected return from being in state $s$, while $Q^\pi$ measures the expected return from taking action $a$ in state $s$. 

In addition to maximizing the cumulative discounted reward, soft actor-critic (SAC) \cite{haarnoja2018soft1} employs a stochastic policy and adds an entropy term of the policy $\mathcal{H}$ to the target (\ref{backgroud-pi}) to facilitate exploration. The entropy term is equivalent to the KL divergence between the agents' policy and a uniformly random action prior. In this paper, we adopt the version that can automatically tune the temperature hyperparameter \cite{haarnoja2018soft}.

\subsection{Learning Control and Training Architecture}
The architecture of the proposed end-to-end driving algorithm is illustrated in Fig. \ref{result-architecture}, consisting of three main phases. 
The first phase entails the specification of objectives and control targets for the autonomous driving system by a designer. 
Subsequently, the RL agent interacts with a driving simulator to discover a control policy that satisfies the predefined objectives, which is represented as a neural network. 
Finally, the control policy undergoes comprehensive validation for rare and hazardous traffic scenarios in the simulator before it can be safely deployed in the real world.

During the first phase, the control objectives are defined by a range of desired driving behaviors, such as driving at high speeds. These objectives are consolidated into a reward function that assesses the quality of the state at each time step while penalizing the control policy for arriving at undesirable terminal states, such as crashes and running out of road.

The second phase employs a high-performance RL algorithm that collects data and uses the provided driving simulator environment to identify a control policy that maximizes the reward function. The collected simulator data enables the RL algorithm to find a near-optimal policy that meets the specified control objectives.

In the third phase, the control policy is deployed in the target vehicle, generating control commands based on the real-time traffic environment and predetermined control targets. This decentralized, end-to-end control algorithm can handle complex and dynamic driving scenarios with varying degrees of autonomy without relying on predefined rules, minimizing dependencies and eliminating unnecessary computations, compared to the traditional centralized and optimization-based control algorithm.

\subsection{Implementation Details}
\subsubsection{State Space}
In this study, we employ a state representation technique that utilizes a state vector to encode features of the observable state space. The state vector can be partitioned into four distinct components, each capturing different features of the observable state space. The first component, the state of the target vehicle, serves as the primary feature of the ego-vehicle and captures motion-related data such as steering, heading, velocity, and relative distance to boundaries. The second component, the navigation information, facilitates the movement of the target vehicle towards its intended destination by computing a route from the spawn point to the destination and generating a set of checkpoints along the route separated by predetermined intervals. The relative distance and direction to the next checkpoint from its current position are then given as the navigation information.

The third component comprises a 240-dimensional vector that characterizes the environment surrounding the target vehicle using LiDAR-like point clouds. The LiDAR sensor scans the surrounding environment using 240 lasers within a horizontal field of view of 360 degrees and a maximum detecting radius of 50 meters from the target vehicle, with a horizontal resolution of 1.5 degrees. Each vector entry reflects the relative distance of the nearest obstacle within a specified direction and is a real number ranging between 0 and 1, subject to Gaussian noise.

The last component, the state of surrounding vehicles, is obtained via V2X communication. In the setting of this paper, each vehicle is equipped with wireless communicators such as Dedicated Short Range Communication (DSRC) and cellular devices to support vehicular communications. This allows the target vehicle to access anonymous motion-related data of the four closest surrounding vehicles.

In summary, the proposed state representation technique partitions the state vector into distinct components, each capturing a unique set of features of the observable state space. By doing so, the technique facilitates an efficient and accurate representation of the state space, thereby improving the performance of the proposed AVs control system.

\subsubsection{Action Space}
This paper employs two normalized actions as input to control the lateral and longitudinal motion of the target vehicle, i.e., $\textbf{a} = [a_1,a_2]^T\in (0,1)$. These normalized actions can be subsequently converted into low-level continuous control commands, namely steering $u_s$, acceleration $u_a$, and brake signal $u_b$ in the following ways:
\begin{subequations}\label{action}
		\begin{align}
			&u_s=S_\text{max} a_1,\\
			&u_a=F_\text{max} \max\{0,a_2\},\\	
			&u_b=-B_\text{max} \min\{0,a_2\},
		\end{align}	
	\end{subequations}
\noindent where $S_\text{max}$ denoted the maximum steering angle, $F_\text{max}$ the maximum engine force, and $B_\text{max}$ the maximum brake force.

\subsubsection{Reward Function}
The reward function $R$ is composed of three parts as follows:
\begin{equation}\label{reward function 1}
R = R_{\text{term}} + (c_1 R_{\text{speed}} + c_2 R_{\text{disp}}) - ( R_{\text{crash}} + R_{\text{out}}).
\end{equation}
\noindent The first term $R_{\text{term}}$ denotes the principal reward, which is non-zero only when the target vehicle reaches the destination. 
The second term is consisted of two auxiliary rewards, namely the speed reward $R_{\text{speed}}$ and the displacement reward $R_{\text{disp}}$, which provide dense signals to stimulate the agent to speed up and move forward, respectively. 
Specifically, the speed reward $R_{\text{speed}}=v_t/v_{\text{max}}$, where the $v_t$ and $v_{\text{max}}$ denote the current and maximum velocity, respectively.
Additionally, the displacement reward $R_{\text{disp}} = d_t-d_{t-1}$, where the $d_t$ and $d_{t-1}$ denote the longitudinal movement of the target vehicle in the Frenet coordinates of the current lane between two consecutive time steps.
Apart from this, the third term is referred to as the punishment function. Considering that collisions cannot be avoided at the beginning of training, we will not terminate the episode if the vehicle gets out of road or involves in a collision to improve training efficiency. By contrast, we will generate negative signals at every moment when the target vehicle crashes or run out off the road.

However, we observe that the RL agent cannot achieve acceptable performance with the reward function (\ref{reward function 1}). For example, it is observed that if the penalty is too low, the agent may rush straight to the destination without following the lane. Besides, if the penalty is too high, the agent may not dare to go forward but stop at the respawn point to escape from punishment.
We found a useful trick to avoid such undesired driving behaviors: modifying the positive reward for being closer to the destination to a penalty, whose value decreases with the distance from the destination. The reshaped reward function $R'$ can be calculated as follows:

\begin{equation}\label{reward function 2}
R' = R_{\text{term}} + c_1 R_{\text{speed}} - (c_2 R_{\text{disp}} + R_{\text{crash}} + R_{\text{out}}).
\end{equation}

\linespread{1.3}
\begin{table}[t]
\centering
\caption{Parameters setting for simulations}
\label{parameters setting for simulations}
{\begin{tabular}{|l|l|}
\hline
\text{Description}   & \text{Notation and Value} \\ \hline
Simulation Time-Step &  $\Delta t = 0.02\, s$ \\ \hline
Control Frequency & $f=50\,HZ$ \\ \hline
Maximum Episode length & $T=1000$ \\ \hline
Number of Lanes  & $3$ \\ \hline
Width of Lane  & $3.5\, m$ \\ \hline
Length of Entrance Segment  & $50\, m$ \\ \hline
Speed Limit  & $80\, km/h$ \\ \hline
Traffic Density  & $0.1\, \text{veh}/10m$ \\ \hline
\end{tabular}}
\end{table}

\subsection{Multi-Task Learning Over Configurations}
This study considers multiple configurations with varying vehicle densities and traffic scenarios. However, training a separate policy for each configuration would be inefficient and time-consuming. To address this issue, we initialize separate environments for each configuration in the configuration settings during training. Our RL algorithm receives trajectories from all environments and batches the gradient update due to these trajectories at each training step.
By employing multi-task learning, we can achieve efficient and effective training of the policy, which can generalize well across a range of vehicular system configurations. The proposed method allows for a detailed exploration of the configuration space, and it facilitates policy learning for different scenarios while avoiding the high costs associated with training separate policies for each configuration.

\section{Experiment Results and Discussion}
\subsection{Experimental Setup}
\linespread{1.3}
\begin{table}[t]
\centering
\caption{Hyperparameters}
\label{parameters setting for training}
{\begin{tabular}{|l|l|}
\hline
\text{Description}   & \text{Value} \\ \hline
Learning Rate &  $10^{-4}$ \\ \hline
Discount Factor &  $\gamma = 0.99 $ \\ \hline
Scaling Factor & $5$ \\ \hline
Hidden Layer & $512 \times 256 \times 256 \times 64$ \\ \hline 
Buffer Size & $ 10^{6} $ \\ \hline
Batch Size & $ 256 $ \\ \hline
Decision Repeat & $ 10 $ \\ \hline
Training Frequency & $ 2 $ \\ \hline
Training Timestep & $ 10^{6} $ \\ \hline
\end{tabular}}
\end{table}

In order to facilitate comprehensive validations for rare and dangerous traffic scenarios before the autonomous driving policy can be safely deployed in the real world, this paper trains and evaluates the proposed DRL algorithm in the open-source MetaDrive simulation platform \cite{li2022metadrive}. Despite various driving simulation platforms available for algorithm training \cite{xu2021opencda,hallyburton2022avstack}, we prefer MetaDrive due to its ability to generate an infinite number of diverse driving scenarios, which supports the research of generalizable DRL algorithms. 

In order to cover as diverse driving occasions as possible, we generate 100 traffic scenarios using different random seeds at the unsignalized intersection presented in Fig. \ref{scenarios-intersection}.
The initial conditions of the training scenarios (e.g., vehicle types, initial locations, and desired routes) are randomly configured by the simulator at the beginning of each episode. 
Additionally, in the setting of this paper, HDVs are assigned to random spawn points on the map and working in a trigger mode to maximize the interaction between target vehicles and traffic vehicles while saving computing resources. In other words, they will stay still in the spawn points until the target agent (i.e., the AV) enters their trigger zone. 
The other parameter settings for simulation are listed in Table \ref{parameters setting for simulations}.

In this paper, the actor and critic networks share the same structure, which consists of four fully connected layers to generate the desired control commands. 
The empirically best-performing hyperparameters and network architecture are listed in Table \ref{parameters setting for training}. We conduct experiments on MetaDrive \cite{li2022metadrive} with RL algorithms implemented in Stable Baseline3 \cite{raffin2021stable}. All the networks are trained with Adam optimizer in PyTorch using an NVIDIA GeForce GTX 960M GPU.

\subsection{Comparison Baselines}
This study compares the proposed $\mathtt{V2X-Lead}$ with three benchmarking algorithms, which are described as follows.
First, $\mathtt{SAC}$ is a popular off-policy RL algorithm that has shown promising results in various applications. In this study, we chose SAC as a baseline algorithm to train the RL agent using the reward function (\ref{reward function 1}).
Second, $\mathtt{SAC-RC}$ is a reward-shaping variant of the classical SAC algorithm, which trains the RL agent using the reward function (\ref{reward function 2}).
Third, $\mathtt{Lead}$ is an ablated version of our proposed V2X-Lead method. It retains the reshaped reward function used in SAC-RS but removes the vehicular communication component used in V2X-Lead. As a result, Lead does not take advantage of the V2X communication data to make decisions, and the RL agent can only rely on its local observation to learn optimal policies.

\subsection{Evaluation Metrics}
In this paper, two indicators are proposed to evaluate the performance of a given driving agent, namely, the success rate and completion time. The success rate is defined as the ratio of episodes where the agent reaches the destination without collisions, measuring the safety performance of the driving agent. The success rate is also considered a suitable measurement for evaluating generalization because it provides a more consistent evaluation across diverse scenes with different road lengths and traffic densities, compared to the cumulative reward, which can vary drastically.

The completion time measures the efficiency of the driving behaviors, which is calculated as the average time cost for successful trials. A shorter completion time indicates better traffic efficiency and the ability to avoid congestion as long as safety is ensured. It is worth noting that the completion time is only calculated for successful trials since the agent's performance during unsuccessful trials is irrelevant to its efficiency.

\linespread{1.3}
\begin{table}[t]
\caption{Evaluation Results}
\label{tab:my-table2}
\resizebox{\columnwidth}{!}{%
\begin{threeparttable}
\begin{tabular}{rcccccc}
\toprule
\multirow{2}{*}{\makecell[c]{Comparison\\Algorithms}} & \multicolumn{2}{c}{$\mathtt{T-Int}$} & \multicolumn{2}{c}{$\mathtt{4-Way-Int}$} & \multicolumn{2}{c}{$\mathtt{Roudabout}$} \\ \cmidrule(r){2-3}\cmidrule(r){4-5}\cmidrule(r){6-7}
                                    & SR $\uparrow$  & CT $\downarrow$ & SR $\uparrow$  & CT $\downarrow$  & SR $\uparrow$  & CT $\downarrow$  \\ \midrule
\multicolumn{1}{l}{Regular Traffic}   & ($\%$)   & ($s$)    & ($\%$)   & ($s$)    & ($\%$)   & ($s$)                                                      \\ \hline
$\mathtt{SAC}$                                 & $82.4$ & $15.4$ & $78.3$ & $15.6$ & $62.0$ & $21.7$ \\
$\mathtt{SAC-RS}$                                 & $87.4$ & $\textbf{11.7}$ & $84.2$ & $12.8$ & $66.3$ & $\textbf{20.8}$ \\
$\mathtt{Lead}$                            & $98.6$ & $19.1$ & $95.6$ & $20.1$ & $\textbf{84.5}$ & $37.1$ \\ \rowcolor{Gray}
\textbf{$\mathtt{V2X-Lead}$}                            & $\textbf{99.7}$ & $11.8$ & $\textbf{99.3}$ & $\textbf{12.6}$ & $84.3$ & $32.4$ \\
\midrule
\multicolumn{7}{l}{Dense Traffic}     \\ \hline
$\mathtt{SAC}$                                 & $  78.9$ & $17.2$ & $74.7$ & $18.7$ & $40.3$ & $\textbf{22.9}$ \\
$\mathtt{SAC-RS}$                                   & $86.8$ & $13.2$ & $78.0$ & $13.4$ & $53.6$ & $25.6$ \\
$\mathtt{Lead}$                                   & $97.0$ & $19.9$ & $95.7$ & $22.9$ & $72.4$ & $26.5$ \\ \rowcolor{Gray}
\textbf{$\mathtt{V2X-Lead}$}                               & $\textbf{97.3}$ & $\textbf{12.4}$ & $\textbf{96.6}$ & $\textbf{12.9}$ & $\textbf{74.7}$ & $25.0$ \\ 
 \bottomrule
\end{tabular}%
  \begin{tablenotes}
  \linespread{1.1}
        \footnotesize
        \linespread{1}
        \item[1] Abbr.: T-Intersection ($\mathtt{T-Int}$) , 4-Way-Intersection ($\mathtt{4-Way-Int}$), Average Success Rate (SR), Average Completion Time (CT).
        \item[2] 500 different traffic scenarios for each configuration were generated using new random seeds, which are different from the situation in the training phase.
      \end{tablenotes} 
    \end{threeparttable}}
\end{table}

\begin{figure}[b]
\centering
\includegraphics[width=0.48\textwidth]{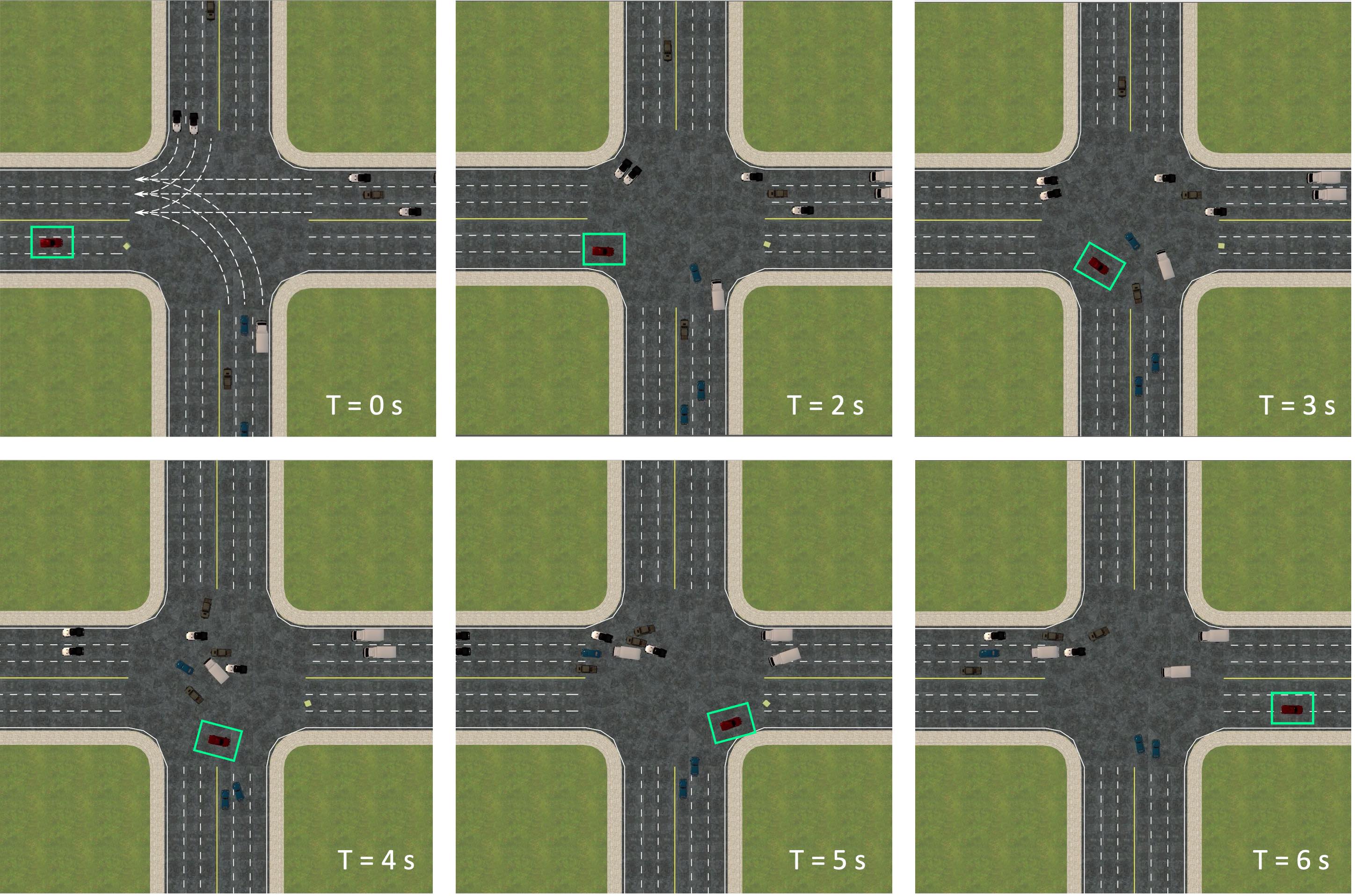}
\caption{Qualitative demonstration of the flexibility of the trained agent.}
\label{2}
\end{figure}

\subsection{Performance Evaluation}
The section discusses the evaluation of the proposed $\mathtt{V2X-Lead}$ approach and its comparison with three benchmark algorithms under different traffic scenarios without signal controllers. The evaluation results are presented in Table \ref{tab:my-table2}. While the T-intersection and 4-way-intersection scenarios are categorized as seen, the roundabout is a new scenario that can test the generalization ability of the proposed algorithm.

\subsubsection{Quantitative Evaluation in Seen Scenarios}
The study found that driving across the 4-way-intersection is more challenging than the T-intersection, as evidenced by the lower success rates of all RL algorithms at 4-way-intersection in all tested scenarios. Additionally, the success rate exhibited a downward trend while the completion time showed an upward trend as the traffic density increased, owing to the increased times of interactions with other vehicles accomplished by a higher possibility of collisions.

Further analysis revealed that the safety performance of $\mathtt{SAC-RS}$ was better than the baseline RL algorithm, $\mathtt{SAC}$, in all tested scenarios, indicating the effectiveness of the reward engineering conducted. The inclusion of the speed reward $R_{\text{speed}}$ and the displacement reward $R_{\text{disp}}$ facilitated the occurrence of efficient driving behaviors and resulted in a second-ranked completion time indicator, only behind $\mathtt{V2X-Lead}$.

Additionally, the safety performance of $\mathtt{Lead}$ showed substantial growth compared to $\mathtt{SAC}$ and $\mathtt{SAC-RS}$ in all tested scenarios. It exhibited an average increment of 23.2\% and 15.2\% in the success rate indicator at the T-intersection and 4-way-intersection, respectively. This is owing to its well-designed training pipeline and hyperparameters that accommodated well with the setting of unsignalized intersections. However, the improved safety performance of $\mathtt{Lead}$ came at the expense of traffic efficiency, leading to conservative driving behavior and the longest completion time among the comparison algorithms.

Finally, the proposed $\mathtt{V2X-Lead}$ approach showed the best performance in terms of safety and efficiency. It maintained success rates of over 99\% and 96\% in regular and dense traffic, respectively. Compared to $\mathtt{Lead}$, $\mathtt{V2X-Lead}$ improved efficiency while ensuring safety by leveraging the advantages of V2X communication to identify the states of surrounding vehicles even if they were blocked by other vehicles, which the LiDAR sensor could not detect.

\linespread{1.1}
\begin{figure*}[t]
\centering
\includegraphics[width=0.85\textwidth]{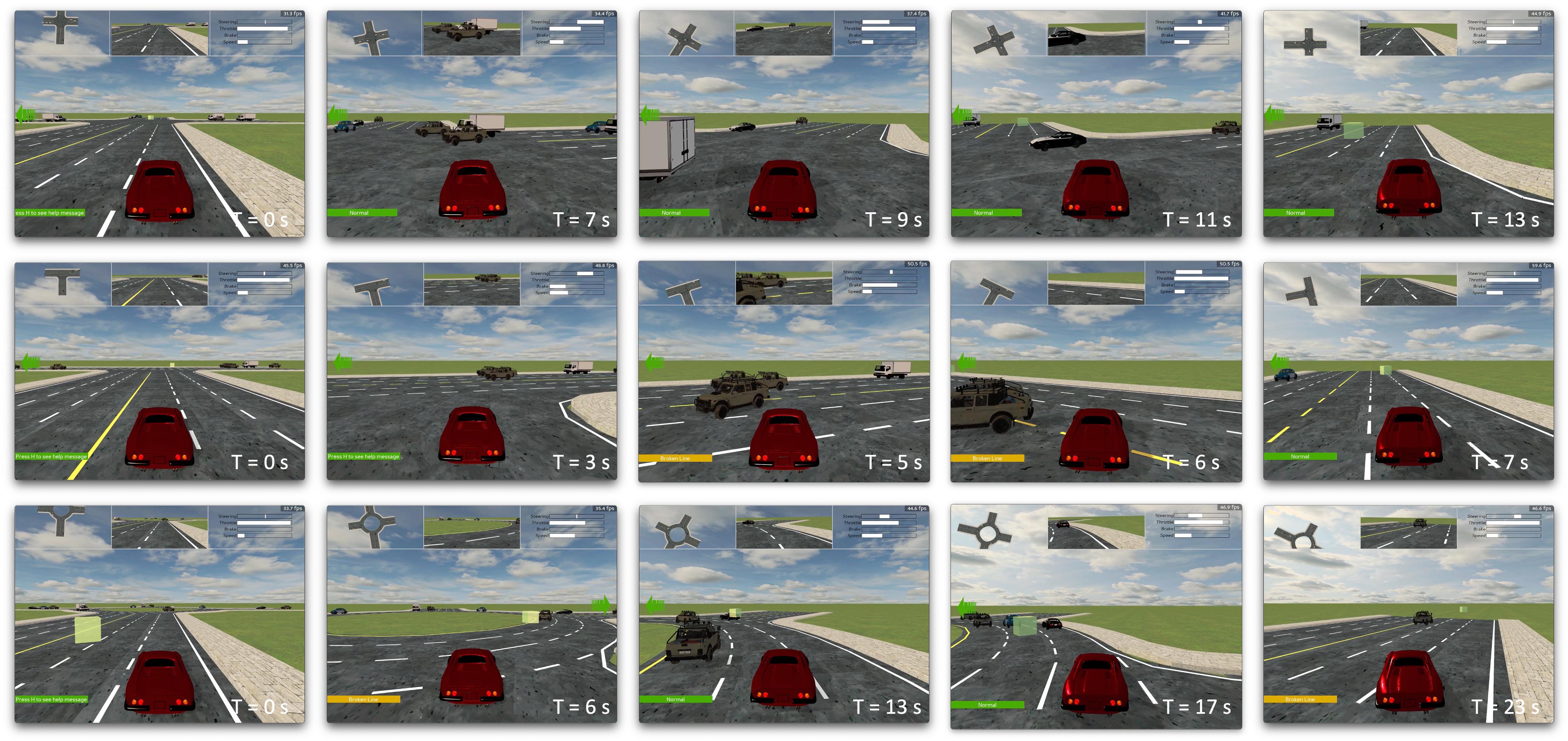}
\caption{Qualitative evaluation results of V2X-Lead in different driving scenarios. The top, middle, and bottom rows demonstrate the driving scenarios at the 4-way-intersection, T-intersection, and roundabout, respectively. The target vehicle colored in red is controlled by the RL agent, while the other social vehicles are driven by the HDV model. The time stamps for each image are shown in white text. More demonstration videos are available at {\tt\small http://zhiyun-cn.github.io/}.
}
\label{4}
\end{figure*}

\subsubsection{Quantitative Evaluation in Unseen Scenarios}
Regarding the Roundabout scenario, the experiment yielded noteworthy results. First, it was observed that all algorithms demonstrated a decline in their performance indicators as the testing scenarios shifted from unsignalized intersections to roundabouts. However, $\mathtt{SAC}$ algorithm was particularly affected, exhibiting a significant 46\% degradation in the success rate indicator under high-density traffic conditions.
Second, $\mathtt{SAC}$ and $\mathtt{SAC-RS}$ demonstrated the shortest completion time. Nevertheless, their success rates were unsatisfactory, which could be attributed to reckless driving behavior. Specifically, the agents trained by these algorithms failed to apply the driving skills they had learned at intersections to other scenarios.
Third, despite $\mathtt{Lead}$ and $\mathtt{V2X-Lead}$ algorithms requiring more time to pass through the roundabout, their success rate degradation percentages remained at 15\% and 23\% in regular and dense traffic, respectively. This finding suggests that the proposed multi-task learning approach generalizes better than the benchmark algorithms.

\subsubsection{Qualitative Evaluation}
To provide readers with a comprehensive understanding of our proposed method, we present qualitative results in Fig. \ref{4}, which showcases the agent's performance in various traffic scenarios. In the first scenario, the agent attempts to make an unprotected left turn at a 4-way intersection but encounters a stream of vehicles blocking its path when it crosses the stop line at $t=7 \,s$. To ensure the safety of all road users, the agent decelerates and turns right slightly to identify a suitable time slot to bypass the moving obstacle from its backside. Eventually, the agent accelerates at $t=11 \,s$, maneuvering through the gap between two vehicles from other lanes and successfully arriving at their destination.
Likewise, in the second scenario at a T-intersection, the agent slows down as it approaches the stop line ($t=3 \,s$) and finds an appropriate time slot to bypass a moving obstacle from its backside at $t=6 \,s$. By applying this lane-changing strategy, the agent successfully reaches its intended destination.
Lastly, in the third scenario at a roundabout, the agent successfully merges into the target lane by turning right at $t=6 \,s$ and then maintains its position using a lane-keeping maneuver until encountering a slowly moving vehicle at $t=13 \,s$. In response, the agent swiftly changes its lane to overtake the obstacle and continue towards its destination.
These results provide compelling evidence of the effectiveness and safety of our proposed method in navigating challenging traffic scenarios.

\subsection{Discussion}

It can be seen from Fig. \ref{2} that the agent has learned some tricks to cross the intersection efficiently without collisions. For example, if the target vehicle is required to go straight at the intersection, it will not simply stay before the stop line to give way. By contrast, it will change the original straight line and find a new curved path to cross the traffic in an appropriate time.
This phenomenon happens because SAC adds an entropy term in the objective function and aims to maximize expected return and entropy simultaneously. In other words, SAC wants to succeed at the task while acting as randomly as possible; thus, diversified driving behaviors are likely to emerge.

\section{Conclusion}

In this work, we propose an innovative LiDAR-based end-to-end autonomous driving method with V2X communication integration, termed V2X-Lead. The proposed method integrates multiple data sources, including but not limited to the LiDAR sensor and V2X communication, through data fusion and employs a model-free and off-policy RL algorithm with a well-designed reward function to train the driving agent. The multi-task learning technique is utilized to enhance the generalization ability of the RL agent across various driving tasks and scenarios. The experimental results show that the proposed method effectively improves safety and efficiency in the task of traversing unsignalized intersections under mixed-autonomy traffic conditions and exhibits generalizability to previously unseen environments, such as roundabouts, outperforming benchmarking methods.

This study opens several directions for future work. 
First, we aim to generalize the proposed algorithm further and develop a unified control methodology that can deliver robust performance across a variety of traffic scenarios. 
Second, we will employ multi-agent RL to enhance the interaction of the target vehicle under control with other AVs and HDVs. 
Third, we plan to inject additional randomization and stochasticity in various aspects of the simulation to facilitate the learning of robust policies for simulation-to-reality transfer, ensuring the safe deployment of the proposed algorithm in the real world. 
Finally, we aim to leverage the interactive protocol proposed by \cite{tsao2022trust} to establish a verifiable communication scheme between authorized traffic infrastructures and vehicles, ensuring the authenticity, integrity, and confidentiality of the mobility data and control commands being transmitted and reducing the risk of cybersecurity issues.
\linespread{1.0}

\bibliography{bibRef.bib}

\end{document}